\documentclass{article}

\usepackage{microtype}
\usepackage{graphicx}
\usepackage{subcaption}
\usepackage{booktabs}
\usepackage{multirow}
\usepackage{xcolor}
\usepackage{pifont}
\usepackage{xspace}
\usepackage{tikz}
\usetikzlibrary{positioning,arrows.meta,shapes.geometric,fit,backgrounds,calc}
\usepackage{algorithm}
\usepackage{algorithmic}
\usepackage{xurl}
\usepackage[hyperfootnotes=false]{hyperref}
\usepackage[most]{tcolorbox}

\newtcblisting{promptbox}[1]{%
  enhanced,
  breakable,
  listing only,
  listing options={
    basicstyle=\ttfamily\scriptsize,
    breaklines=true,
    breakatwhitespace=true,
    columns=fullflexible,
    keepspaces=true,
    showstringspaces=false
  },
  colback=white,
  colframe=black!35,
  colbacktitle=black!7,
  coltitle=black,
  fonttitle=\bfseries\footnotesize,
  title={#1},
  boxrule=0.3pt,
  arc=1pt,
  left=4pt,
  right=4pt,
  top=3pt,
  bottom=3pt,
}

\newcommand{\cmark}{\textcolor{green!55!black}{\ding{51}}}
\newcommand{\xmark}{\textcolor{red!70!black}{\ding{55}}}
\newcommand{\tmark}{\textcolor{orange!90!black}{$\triangle$}}

\newcommand{\method}{\emph{Online Agent-as-a-Judge}\xspace}

\usepackage[accepted]{icml2026}

\usepackage{amsmath}
\usepackage{amssymb}
\usepackage{mathtools}
\usepackage{amsthm}

\usepackage[capitalize,noabbrev]{cleveref}

\theoremstyle{plain}

\theoremstyle{definition}

\theoremstyle{remark}

\usepackage[disable,textsize=tiny]{todonotes}

\icmltitlerunning{\emph{\smash{Online Agent-as-a-Judge}}: Situation-Generating Evaluation for Interactive Agents}

\begin{document}

\twocolumn[
  \icmltitle{\method: Situation-Generating Evaluation for Interactive Agents}

  \icmlsetsymbol{equal}{*}
  \begin{icmlauthorlist}
    \icmlauthor{Hyogon Ryu}{anon}
    \icmlauthor{Jeonghwan Kim}{anon}
    \icmlauthor{Yewon Lim}{anon}
    \icmlauthor{Chaeun Lee}{anon}
    \icmlauthor{Jeongwook Kim}{anon}
    \icmlauthor{Donghoon Ham}{anon}
  \end{icmlauthorlist}
  \icmlaffiliation{anon}{KRAFTON}
  \icmlcorrespondingauthor{Hyogon Ryu, Donghoon Ham}{hyogon.ryu, donghoon.ham@krafton.com}
  \icmlkeywords{LLM Agents, Agent-as-a-Judge, Game Agents, Life Simulation, Evaluation}
  \vskip 0.3in
]

\printAffiliationsAndNotice{}

\begin{abstract} Evaluating LLM-powered interactive social agents is challenging because socially relevant behaviors depend not only on isolated outputs, but also on prior interactions, social roles, and downstream actions. Existing methods typically allow a target agent to act freely in an environment and then score the resulting trajectory. However, this passive setup can miss capabilities that only become observable under specific social circumstances; for example, conflict handling may remain untested if no disagreement arises. We propose \textbf{\method}, a situation- generating evaluation framework for interactive social agents. \method deploys an in-world evaluator agent that interacts with the target agent through the environment's native dialogue and action protocol, actively eliciting situations relevant to the evaluation criteria. The resulting trajectories provide evidence for assessing both immediate responses and subsequent behavior. In a life-simulation environment with $32$ designer-authored social criteria, \method improves criteria coverage and agreement with human labels, yielding more reliable evidence- grounded evaluations of behaviors that passive methods can leave unobserved. \end{abstract}
\section{Introduction}
\label{sec:intro}

Large language models (LLMs) are increasingly used as the cognitive core of interactive agents that observe state, communicate, and act through structured environment protocols~\citep{liu2023agentbench,xie2024osworld,zhou2023webarena,wang2023voyager}. This shift is especially visible in social simulations. Generative Agents showed how LLM-driven characters can inhabit an interactive sandbox and exhibit believable individual behaviors and emergent social dynamics~\citep{park2023generative}. Concordia extends this direction toward generative agent-based modeling, where agents describe intended actions in natural language and a Game Master resolves them into grounded outcomes~\citep{velez2024concordia}. LLM agents are becoming participants in rich social worlds rather than isolated response generators.

This raises a difficult evaluation question. In persistent social environments, success is rarely captured by a single terminal outcome; it is a pattern of context-sensitive behavior: whether an agent maintains a stable role, remembers prior interactions, follows through on commitments, and repairs social mistakes. Many of these behaviors are precisely the ones that ordinary interaction traces do not contain, such as a broken promise, a refused request, an awkward social ask, or a moment of distress that requires emotional support. The evaluation bottleneck is therefore not only how to judge a trace, but how to obtain a trace that contains the situation of interest in the first place.

Existing evaluation methods only partially address this bottleneck. Many LLM-as-a-judge methods score completed outputs or pre-recorded trajectories~\citep{zheng2023llmjudge,liu2023geval,zhuge2025agentjudge,shi2026ajbench}, which is effective when relevant evidence is already present but provides little mechanism for producing such evidence when it is absent. Even evaluators that retrieve additional memory or world-state information~\citep{gou2025mind2web2agenticsearch,lu2025agentrewardbench} primarily expand access to existing evidence rather than create the situations needed to test a criterion. What is missing is an evaluator that can intervene during evaluation: one that interacts with the target to elicit relevant social situations, much like a human playtester placing characters in difficult situations to observe their responses.

\begin{table*}[t]
\caption{Comparison of evaluation paradigms for interactive social agents.
\cmark{} = supported, \tmark{} = partially supported, \xmark{} = not supported.}
\label{tab:methods}
\centering
\footnotesize
\setlength{\tabcolsep}{5.5pt}
\renewcommand{\arraystretch}{1.12}
\begin{tabular}{@{}lccccc@{}}
\toprule
Evaluation paradigm & Process & Active & Context & Custom & Scalable \\
\midrule
Outcome benchmark & \xmark & \xmark & \xmark & \xmark & \cmark \\
Offline LLM-as-a-Judge & \tmark & \xmark & \tmark & \tmark & \cmark \\
Offline Agent-as-a-Judge & \cmark & \xmark & \cmark & \tmark & \cmark \\
Designer playtest & \cmark & \cmark & \cmark & \cmark & \xmark \\
\textbf{\emph{Online Agent-as-a-Judge}} & \cmark & \cmark & \cmark & \cmark & \cmark \\
\bottomrule
\end{tabular}

\vspace{2pt}
\begin{minipage}{0.92\linewidth}
\footnotesize
\emph{Process} = evaluates behavior over interaction, not only final outputs.
\emph{Active} = participates in the environment via its native action/dialogue protocol to elicit criterion-relevant situations.
\emph{Context} = accesses world, memory, and relationship state.
\emph{Custom} = supports designer-authored behavioral criteria.
\emph{Scalable} = runs repeatedly without manual playtesting effort.
\end{minipage}
\end{table*}

We propose \method, a situation-generating evaluation framework for interactive social agents. Rather than evaluating only after an interaction has completed, \method embeds an online judge, in the same environment as the target during evaluation. The online judge is assigned an in-world role and persona, allowing it to interact with the target as another character through the environment's native dialogue and action interfaces. The online judge is additionally equipped with evaluation-specific read-only tools that expose criterion-relevant context, such as other agents' roles, relationships, memories, and recent interactions, so it can decide which situation to elicit. Given a designer-authored behavioral criterion, the online judge plans a probe, elicits the situation through native interaction, and observes the target's response and any follow-through. The resulting trajectory is then used to produce an evidence-grounded judgment.

We evaluate \method in a life-simulation environment using a five-character family scenario and designer-authored criteria covering role consistency, memory continuity, coordination, emotional support, and conflict handling. We compare \method against two offline baselines that cannot intervene in the environment: an offline LLM-as-a-judge, which evaluates fixed interaction logs, and an offline agent-as-a-judge, which retrieves criterion-relevant context from those logs before evaluation. We report evidence coverage and agreement with per-criterion human labels.

This paper makes three contributions. First, we identify \emph{situation availability} as a central bottleneck in evaluating interactive social agents: relevant evidence may be missing because the appropriate situation never arises. Second, we propose \method, a framework in which an online judge interacts with the target during evaluation to elicit criterion-relevant situations through native dialogue and action interfaces. Third, we instantiate \method in a life-simulation with $32$ designer-authored criteria across eight social domains, showing that active elicitation improves evidence coverage and agreement with per-criterion human labels, especially for conflict handling and emotional support.

\begin{figure*}[t]
\centering
\includegraphics[width=\linewidth]{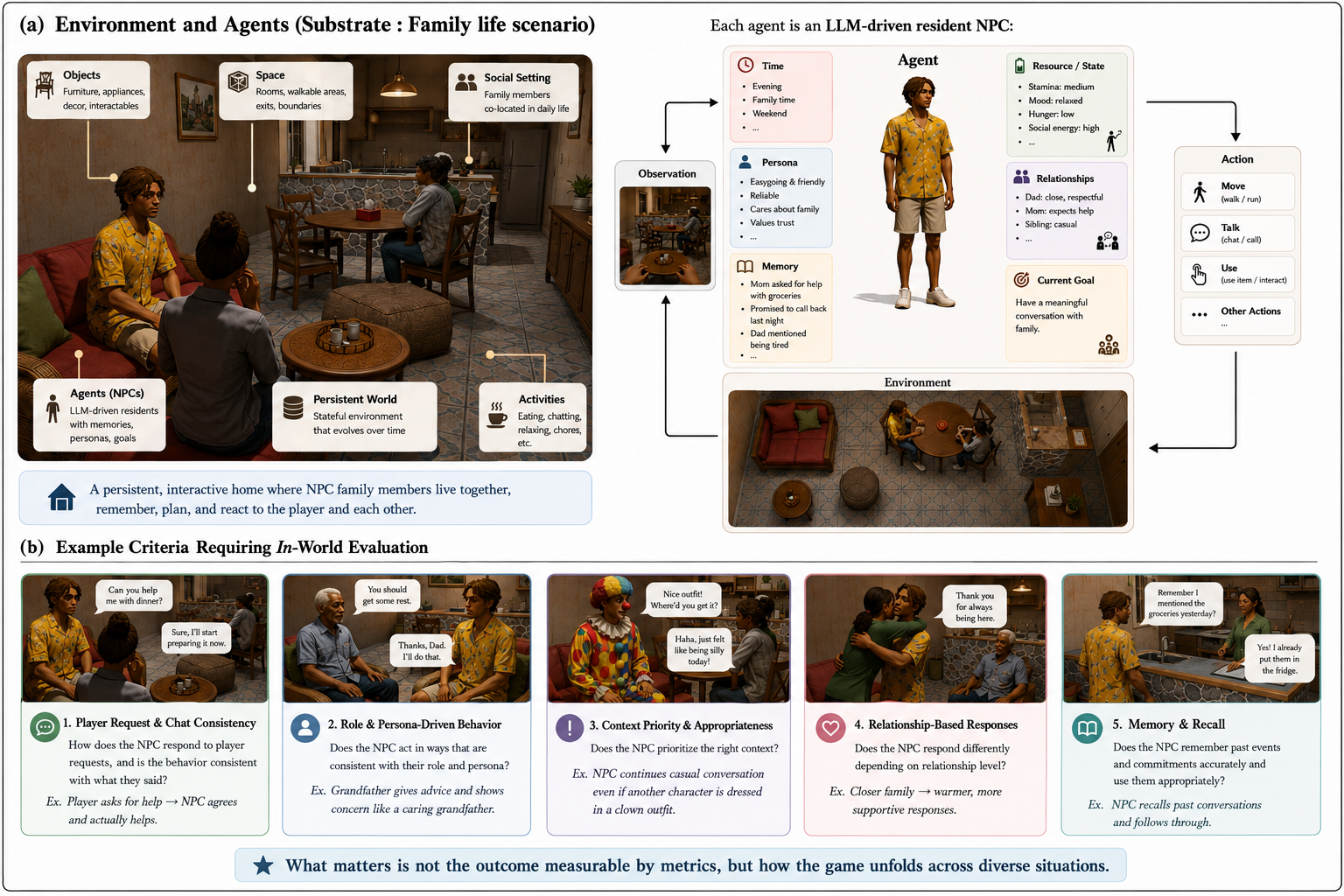}
\caption{\textbf{Life simulation as an evaluation target.} (a) The world is a persistent home with multiple NPCs; what matters is not a single outcome but how the target agent (green) handles a stream of small social situations. (b) Each agent runs an observe--plan--act loop over the world's structured protocol; an online judge participates in the \emph{same} loop as one of the NPCs.}
\label{fig:lifesim}
\end{figure*}

\section{Related Work}
\label{sec:related}

\paragraph{LLM-as-a-judge.}
LLM-as-a-judge methods use another LLM to score model outputs such as chat replies, summaries, or code~\citep{zheng2023llmjudge,liu2023geval,zhou2025personaeval,lee2025correctlyreportllmasajudgeevaluations}. They are convenient when the artifact already exists, but they cannot ask the system under test for a specific behavior, and so cannot evaluate criteria that ordinary outputs do not exhibit.

\paragraph{Agent-as-a-judge and trajectory evaluation.}
Agent-as-a-Judge generalises this idea by giving the judge tools to inspect environment state and intermediate steps~\citep{zhuge2025agentjudge,shi2026ajbench,gou2025mind2web2agenticsearch,lu2025agentrewardbench,li2026atbench}. These judges, however, still observe a \emph{completed} interaction or interact with an environment that is separate from the agent under test's runtime environment. They are environment-aware but not situation-generating: they cannot create the social situation a designer wants to probe. \method differs along both axes: it evaluates the target during interaction and actively constructs situations for criteria that ordinary trajectories never reach.

\paragraph{Game and agent benchmarks.}
A growing line of work benchmarks LLM agents in games and interactive environments: ORAK across many video games~\citep{park2025orak}, GVGAI-LLM with infinite procedural games~\citep{li2025gvgai}, VideoGameBench for vision-language game completion~\citep{zhang2025videogamebench}, RPGBench for role-play engines~\citep{yu2025rpgbench}, MineDojo and Voyager for open-ended embodied agents~\citep{fan2022minedojo,wang2023voyager}, OSWorld and WebArena for tool-use agents~\citep{xie2024osworld,zhou2023webarena}, AgentBench across many environments~\citep{liu2023agentbench}, DeepPlanning for long-horizon planning~\citep{zhang2026deepplanning}, and DeliveryBench for profit-driven action~\citep{mao2025deliverybench}. Most score task completion, reward, or success rate. Life-sim agents are different: there is no single task to complete, success is socially defined, and the relevant behaviors must be \emph{induced} by another social participant.

\paragraph{Generative agents and life simulation.}
Generative Agents~\citep{park2023generative} and Social Simulacra~\citep{park2022socialsimulacra} model believable social behavior with LLMs; recent work pushes life simulation as a generative medium~\citep{li2024unbounded,cheng2025animegamer,duan2026lifesim,wang2024d2a}. Believability has been benchmarked in offline ways~\citep{xiao2023simulatebench}, and recent reviews and studies argue that validation, not raw believability, is the bottleneck for generative social simulation~\citep{larooij2026validationgenerativesocialsimulation,wang2025llmbasedhumansimulationsreliable,wu2025llmbasedsocialsimulationsrequire}. We complement this line: rather than asking whether an agent is believable in the abstract, we ask whether an agent satisfies a designer's specific behavioral criteria, and we use an online judge to elicit and check criterion-relevant behavior.

\paragraph{Process-aware and social evaluation.}
Closest in spirit are Concordia's centralised Game Master, which arbitrates situations and stages scenarios to study a range of emergent social-simulation phenomena~\citep{velez2024concordia}, and recent process-aware audits such as M3-Bench's evaluation of mixed-motive social games~\citep{xie2026m3bench}, AgentRewardBench's audit of trajectory evaluations~\citep{lu2025agentrewardbench}, and Meta-Harness on harness-level optimisation~\citep{lee2026metaharness}. Concordia in particular shares the situation-generating intuition behind \method and has been effective at exposing many social phenomena, but it does so from \emph{outside} the agents as an external arbiter, and its evaluation targets are the emergent phenomena themselves rather than per-character behavioral criteria of the kind a game designer authors for a specific NPC in a shipped build. We share these works' concern that final-state metrics miss social process, and add an evaluator that interacts with the target online and changes which processes get observed.

\section{Method}
\label{sec:method}

\begin{figure*}[t]
\centering
\includegraphics[width=\linewidth]{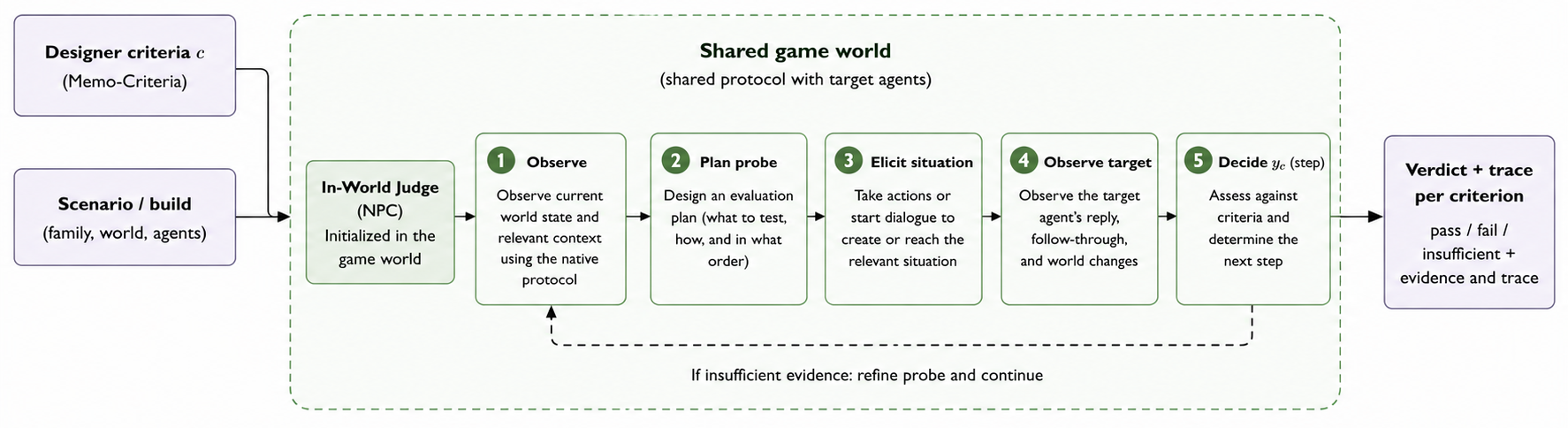}
\caption{\textbf{\method framework.} The judge consumes designer criteria and the current build, plans a probe, enters the same simulation world as the target agent, elicits the relevant situation through dialogue and action, observes the target's reply and follow-through, and either emits a verdict or refines the probe. Because elicitation and observation use the simulator's native protocol, the same evaluation framework survives build changes that would break a fixed benchmark.}
\label{fig:framework}
\end{figure*}
\subsection{Problem setting}

A simulation world $\mathcal{W}$ runs a discrete-time protocol. At each turn, an agent receives a structured observation $o_t \in \mathcal{O}$, including visible characters, dialogue history, scene state, and agent-visible character attributes, and emits an action $a_t \in \mathcal{A}$ that mixes natural-language utterances with simulation-defined actions such as moving, using an object, or proposing an interaction. The \emph{target agent} $\pi_T$ is one such NPC. A designer provides criteria $\mathcal{C}=\{c_1,\dots,c_K\}$ describing soft behavioral properties of $\pi_T$, such as \emph{maintains family role under challenge} or \emph{follows through on a household request}. The full list appears in Appendix~\ref{sec:criteria}. For each criterion $c$, the evaluation assigns a label $y_c \in \{\texttt{pass},\texttt{fail},\texttt{insufficient}\}$ intended to match a designer's judgment.

\subsection{\method}

We instantiate the judge $\pi_J$ as a co-resident participant inside $\mathcal{W}$. The judge intervenes only through the same native dialogue and action protocol as ordinary agents. For evaluation planning, it may also access read-only inspection tools. These tools expose the surrounding scene, nearby characters, recent interactions, and the current evaluation episode, and let the judge update its own probe memory and plan; they cannot modify the world or alter what the target agent observes.

For each criterion, \method maintains an evidence set $E$ of probe episodes collected so far. At each iteration, the judge inspects the state summary $S$ and constructs a probe plan $p$. Executing $p$ through the native simulation protocol yields an episode $\tau$ containing the judge's actions, the target's dialogue responses, any follow-through actions, and relevant observations. We measure $|\tau|$ in environment turns. For exposition, Algorithm~\ref{alg:online_judge} separates the loop into \texttt{Inspect}, \texttt{Plan}, \texttt{Run}, and \texttt{Decide}. In the implementation, inspection, planning, and provisional deciding are carried out by the same probe-loop agent call. That call returns a provisional verdict $y_c$, a confidence $\mathit{conf}\in[0,1]$, and a stop flag. Confidence is logged but does not trigger a separate threshold-based early stop: the judge halts only when the returned stop flag is set under the evidence rules, or when the simulator's wall-clock session budget is exhausted. If the budget is exhausted, a separate scoring step (\texttt{Score} in Algorithm~\ref{alg:online_judge}) renders the verdict from the accumulated evidence. Algorithm~\ref{alg:online_judge} and Figure~\ref{fig:framework} summarize the loop, and \Cref{sec:prompts} describes the four LLM call sites that implement it.

\begin{enumerate}
\item \textbf{Inspect.} Given criterion $c$, the judge queries the evaluation tools to gather criterion-relevant context: who the target is, who is around, what has just happened, and what continuity matters. The information returned is read-only and does not alter the world.
\item \textbf{Plan.} Using the inspected state, the judge produces an evaluation plan for the target chosen at the start of the session: which situation should arise, how to elicit it, and what evidence would constitute pass, fail, or insufficient evidence.
\item \textbf{Elicit.} The judge moves and talks inside $\mathcal{W}$ to construct the situation, such as approaching the target and asking ``can you grab me a coffee?''
\item \textbf{Observe.} The judge records the target's dialogue and \emph{follow-through} actions in subsequent turns, not just the immediate reply.
\item \textbf{Decide.} If the collected non-judge target evidence supports a verdict and the stop rules are met, the judge emits $y_c$. Otherwise it returns to step~1 with a refined probe, such as a different framing, an escalation, or a follow-up that targets the missing evidence.
\end{enumerate}

\begin{algorithm}[t]
\small
\caption{\method evaluation loop for one criterion}
\label{alg:online_judge}
\begin{algorithmic}[1]
\STATE \textbf{Input:} criterion $c$, target $\pi_T$, world $\mathcal{W}$, eval tools $\mathcal{T}$
\STATE \textbf{Output:} $y_c \in \{\texttt{pass},\texttt{fail},\texttt{insufficient}\}$
\STATE $E \gets \emptyset$
\WHILE{simulator session not exhausted}
  \STATE $S \gets \texttt{Inspect}(\mathcal{T}, c, E)$ \COMMENT{query read-only tools}
  \STATE $p \gets \texttt{Plan}(c, E, S)$ \COMMENT{pick next probe}
  \STATE $\tau \gets \texttt{Run}(\mathcal{W}, \pi_J, \pi_T, p)$ \COMMENT{online interaction episode}
  \STATE $E \gets E \cup \{\tau\}$
  \STATE $(y_c, \mathit{conf}, \texttt{stop}) \gets \texttt{Decide}(c, E)$
  \IF{$\texttt{stop}$}
    \STATE \textbf{return} $y_c$
  \ENDIF
\ENDWHILE
\STATE \textbf{return} $\texttt{Score}(c, E)$ \COMMENT{score the accumulated evidence}
\end{algorithmic}
\end{algorithm}

\section{Experiments}
\label{sec:experiments}

\subsection{Implementation Details}

\paragraph{Simulator and scenario.}
We evaluate a life-simulation sandbox with a five-character family, including parent, child, sibling, and grandparent roles. The sandbox supports diverse dialogue and action settings, with characters controlled by either rule-based or LLM-driven policies. In \method{}, one family member is instantiated as the judge, and the remaining family members are target agents under evaluation.

\paragraph{Target agents.}
We evaluate three character backends: random rule-based selection, single-shot LLM prompting, and an observe--think--act agent loop that keeps memory and plans across turns. Each backend is run three times with different random seeds.

\paragraph{Criteria.}
We use $32$ human-curated criteria. \Cref{tab:criteria_domains} summarizes the eight criterion domains; the full criterion list is in Appendix~\ref{sec:criteria}.

\begin{table}[t]
\caption{Criterion domains used in the experiments.}
\label{tab:criteria_domains}
\centering
\small
\begin{tabular}{lc}
\toprule
Domain & \# Criteria \\
\midrule
Conversation/Relationship & 5 \\
Family Role/Persona & 7 \\
Memory/Continuity & 3 \\
Household Coordination & 6 \\
Emotional/Social Support & 4 \\
Agency/Goal Alignment & 2 \\
Play & 2 \\
Conflict/Norm Violation & 3 \\
\midrule
\textbf{Total} & \textbf{32} \\
\bottomrule
\end{tabular}
\end{table}

\paragraph{Comparison baselines.}
Each judge returns one of three labels for each criterion: \texttt{pass}, \texttt{fail}, or \texttt{insufficient}.
\begin{itemize}
\item \textbf{Offline LLM-as-a-Judge.} A one-shot LLM judge that reads the pre-recorded trajectory of a simulation session and emits a verdict for each criterion.
\item \textbf{Offline Agent-as-a-Judge.} An LLM agent that can inspect the trace database, search by keyword, and re-read events, but cannot control the sandbox or create new situations.
\item \textbf{\method{}.} The proposed framework. The judge enters the same simulator session as the target agents, embodies one family member, and creates criterion-relevant situations through the sandbox's native dialogue/action protocol.
\end{itemize}
All three automated judges use GPT-5.4-mini and the same final verdict definitions, so differences in their outputs reflect differences in evidence collection rather than different scoring rules.

\paragraph{Metrics.}
We report two main metrics. First, \emph{criteria coverage} measures the fraction of criteria for which the judge collected enough concrete dialogue/action evidence to emit a \texttt{pass} or \texttt{fail} verdict. Second, \emph{human agreement} measures accuracy against per-criterion human labels.

\subsection{Experimental Results}

\paragraph{Criteria coverage.}
Table~\ref{tab:coverage} reports coverage broken down by criterion domain. \method{} achieves $0.92$ average coverage, compared to $0.56$ for the offline LLM judge and $0.54$ for the offline agent judge: it produces a pass/fail verdict on $92\%$ of cells, while the offline judges return \texttt{insufficient} on roughly half. The coverage gap is large in every domain and largest where the relevant situation rarely appears in passive simulation traces: Conflict/Norm Violation ($1.00$ vs $0.19$/$0.22$) and Emotional/Social Support ($0.89$ vs $0.44$/$0.58$). Conversation/Relationship also shows a meaningful gap ($0.80$ vs $0.67$/$0.60$): offline judges can find some relevant interactions, but still often lack enough evidence because passive sessions do not cover the required range of situations for each criterion.

\begin{table*}[t]
\caption{Criteria coverage by judge and domain. Each cell is the fraction of evaluations with a \texttt{pass} or \texttt{fail} verdict, pooled across three character backends and three random seeds. Domain abbreviations: Cnv = Conversation/Relationship (5 criteria), Fam = Family Role/Persona (7), Mem = Memory/Continuity (3), Hsh = Household Coordination (6), Emo = Emotional/Social Support (4), Agcy = Agency/Goal Alignment (2), Play = Play (2), Cnf = Conflict/Norm Violation (3). Best per column in \textbf{bold}.}
\label{tab:coverage}
\centering
\small
\begin{tabular}{lccccccccc}
\toprule
Judge                          & Avg. & Cnv & Fam & Mem & Hsh & Emo & Agcy & Play & Cnf \\
\midrule
Offline LLM-as-a-Judge         & 0.56 & 0.67 & 0.59 & 0.52 & 0.70 & 0.44 & 0.83 & 0.33 & 0.19 \\
Offline Agent-as-a-Judge       & 0.54 & 0.60 & 0.51 & 0.44 & 0.69 & 0.58 & 0.83 & 0.33 & 0.22 \\
\textbf{\method{}}             & \textbf{0.92} & \textbf{0.80} & \textbf{0.92} & \textbf{0.85} & \textbf{1.00} & \textbf{0.89} & \textbf{1.00} & \textbf{0.94} & \textbf{1.00} \\
\bottomrule
\end{tabular}
\end{table*}

\paragraph{Human agreement.}
Table~\ref{tab:agreement} reports accuracy against human labels. \method{} reaches $0.70$, compared to $0.40$ for the offline agent judge and $0.33$ for the offline LLM judge. As with coverage, the largest gains are in domains where the criterion-relevant situation is socially induced rather than spontaneously displayed: Conflict/Norm Violation ($0.96$ vs $0.17$/$0.17$) and Emotional/Social Support ($0.82$ vs $0.45$/$0.55$). \method{} leads in every domain, although Memory/Continuity remains weak for all judges; we return to this limitation in \Cref{sec:discussion}.

\begin{table*}[t]
\caption{Human agreement, measured as accuracy against per-criterion human labels. Pooling and domain abbreviations are as in Table~\ref{tab:coverage}. Best per column in \textbf{bold}.}
\label{tab:agreement}
\centering
\small
\begin{tabular}{lccccccccc}
\toprule
Judge                          & Avg. & Cnv & Fam & Mem & Hsh & Emo & Agcy & Play & Cnf \\
\midrule
Offline LLM-as-a-Judge         & 0.33 & 0.42 & 0.38 & 0.00 & 0.26 & 0.45 & 0.44 & 0.33 & 0.17 \\
Offline Agent-as-a-Judge       & 0.40 & 0.47 & 0.38 & 0.14 & 0.52 & 0.55 & 0.44 & 0.33 & 0.17 \\
\textbf{\method{}}             & \textbf{0.70} & \textbf{0.58} & \textbf{0.68} & \textbf{0.33} & \textbf{0.76} & \textbf{0.82} & \textbf{0.72} & \textbf{0.73} & \textbf{0.96} \\
\bottomrule
\end{tabular}
\end{table*}

\paragraph{Robustness and class balance.}
\label{sec:robustness}
Table~\ref{tab:class_accuracy} reports pass-label and fail-label accuracy. \method{} correctly identifies $93$ of $114$ human-labelled passes ($82\%$) and $87$ of $144$ human-labelled failures ($60\%$). By contrast, the offline LLM judge identifies $73$ passes ($64\%$) but only $12$ failures ($8\%$), while the offline agent judge identifies $58$ passes ($51\%$) and $46$ failures ($32\%$). Thus the gains in Tables~\ref{tab:coverage} and~\ref{tab:agreement} are not driven by simply predicting \texttt{pass}; \method{} is substantially better at finding true failures.

\begin{table}[t]
\caption{Pass-label and fail-label accuracy against human labels. Non-decisive labels count as non-matches.}
\label{tab:class_accuracy}
\centering
\small
\begin{tabular}{lccc}
\toprule
Judge                          & Pass Acc. & Fail Acc. & Acc. \\
\midrule
Offline LLM                    & 0.64      & 0.08      & 0.33 \\
Offline Agent                  & 0.51      & 0.32      & 0.40 \\
\textbf{Ours}                  & \textbf{0.82} & \textbf{0.60} & \textbf{0.70} \\
\bottomrule
\end{tabular}
\end{table}

\paragraph{Time cost.}
Table~\ref{tab:cost} reports wall-clock time. A designer-run evaluation of one situation per criterion takes roughly $60$ minutes of focused human time. \method{} evaluates the same $32$ criteria in approximately $21$ minutes of wall-clock time by running criteria in parallel.

\begin{table}[t]
\caption{Wall-clock time to evaluate one full $32$-criterion pass against one target-agent configuration.}
\label{tab:cost}
\centering
\small
\begin{tabular}{lc}
\toprule
Method                              & Wall-clock time \\
\midrule
Designer-run evaluation             & $\sim 60$ min        \\
\textbf{\method{}}                  & $\sim 21$ min        \\
\bottomrule
\end{tabular}
\end{table}

\subsection{Analysis}

\paragraph{Where the gap comes from.}
The trajectory-based methods do not lose because they reason worse; they lose because the trace they read often does not contain the relevant situation. The pattern is sharpest in Conflict/Norm Violation and Emotional/Social Support, where criteria depend on triggers such as an insult, a moment of strong distress, or a broken promise. These events rarely arise in ordinary simulation sessions, so offline judges often have nothing concrete to score from. \method{} constructs those triggers on demand. In domains such as Conversation/Relationship and Household Coordination, relevant situations appear more often in passive traces, so the offline judges can sometimes score them and the gap is smaller.

\paragraph{Dialogue vs follow-through.}
Follow-through criteria expose a second limitation of passive traces: the evidence must include both the eliciting request and the later action. A target may verbally accept a household request, then do something unrelated; conversely, it may satisfy the request through action without a clean verbal commitment. Offline trace inspection can judge these cases only when the passive session happens to contain the right initiating situation and enough subsequent action context. \method{} couples the two steps: it creates the request, keeps the evaluation focused on the same target, and observes whether the promised behavior actually occurs.

\paragraph{Where all judges struggle.}
Memory/Continuity is difficult for all judges because the evidence is distributed across long interaction histories rather than concentrated in the probe episode. A correct verdict often requires retrieving the relevant earlier event, deciding whether the target should still remember it, and checking whether the current claim is genuinely grounded in that event rather than merely plausible in context. This turns memory evaluation into a long-context attribution problem: failures can come from the target forgetting, the judge missing the relevant earlier evidence, or the human label depending on a different reconstruction of the prior interaction. The low scores in this domain suggest that memory evaluation needs more targeted probes and better long-context evidence retrieval.

\section{Discussion}
\label{sec:discussion}

\paragraph{What \method is and is not.}
\method is useful when the property of interest is \emph{induced} rather than \emph{displayed}: success is a pattern of context-sensitive behavior, and the situations that reveal it rarely arise in passive logs. It is not a replacement for outcome benchmarks when the target behavior has a well-defined quantitative endpoint. If a social simulation task can be measured by completion time, resource use, score, survival, or another direct quantity, those measures should remain primary; \method is for the behavioral criteria that such quantities miss.

\paragraph{Methodological cautions.}
Two cautions deserve naming. First, an active probe can inadvertently contain the answer it is meant to test: the judge may ask in a way that reveals the expected behavior, supplies the missing context, or performs the criterion-positive action for the target. \method's main defence is the probe gate described in \Cref{sec:prompts}, which rejects probes that make the judge perform the criterion-positive behavior itself or scaffold the desired answer; splitting criteria into dialogue-only and follow-through evidence is a complementary defence, since follow-through is harder to script. Second, the judge's verdicts inherit the biases of the underlying LLM. We mitigate this by separating \texttt{pass}, \texttt{fail}, and \texttt{insufficient} and by sharing the same final-judging policy block across all judges, so that comparisons reflect evidence collection rather than scoring style.

\paragraph{Applicability.}
The framework applies beyond our simulator whenever an evaluator can observe state, enter through the same action protocol as other participants, and create situations that would otherwise be rare. This includes life-simulation games such as \emph{The Sims}~\citep{maxis2000sims} and \emph{inZOI}~\citep{krafton2025inzoi}, long-running conversational assistants, embodied or multi-agent social benchmarks~\citep{xie2026m3bench,zhou2024sotopia}, and role-play or persona evaluation~\citep{zhou2025personaeval,yu2025rpgbench}. Across these settings, the common bottleneck is not scoring a visible outcome but eliciting the situation in which the behavior can be judged.

\section{Conclusion}
\label{sec:conclusion}

In this paper, we proposed \method, a situation-generating evaluation framework for interactive agents in social simulations. By placing a judge inside the target agent's environment and giving it access to the same dialogue and action protocol, \method elicits criterion-relevant situations that rarely appear in passive traces. In a life-simulation build with $32$ designer-authored criteria, \method achieved substantially higher criteria coverage ($0.92$ versus $0.56$ and $0.54$ for the offline LLM-as-a-Judge and offline Agent-as-a-Judge baselines, respectively) and higher human-label agreement accuracy ($0.70$ versus $0.33$ and $0.40$), with the largest gains in domains where the relevant situations rarely arise on their own. These results establish situation generation as a practical path for evaluating socially situated agents whose behavior must be elicited before it can be judged. More broadly, we hope this work highlights active, in-environment evaluation as a methodological tool for studying and validating interactive agents in social simulations.

\section*{Impact Statement}

This paper presents a framework for evaluating interactive agents by actively
eliciting criterion-relevant situations. Potential societal impacts include
improved reliability and safety evaluation of LLM-powered agents in games,
simulations, and other interactive applications. At the same time, active
probing agents could introduce evaluation bias or influence target behavior if
used without safeguards. We discuss mitigations including native-protocol
interaction, probe gating, and judging only non-judge target evidence.

\bibliography{references}
\bibliographystyle{icml2026}

\appendix
\section{Full Criteria Set}
\label{sec:criteria}

The criteria below are the human-curated criteria used in Section~\ref{sec:experiments}. Each criterion records its behavioral question together with its \textbf{form} (general behavioral / situation-specific everyday / situation-specific exceptional) and \textbf{coverage} type (trace-visible / mixed / judge-elicited). Some criteria additionally include positive and negative signal notes. These notes are optional reference guidance, analogous to using a reference-guided judging prompt rather than a plain question-only prompt; they help clarify the intended behavior but are not required for every criterion and are not treated as a separate label.

\subsection*{1. Conversation / Relationship}

\paragraph{C1. Relationship-aware conversation.} \textit{Form: General Behavioral. Coverage: Trace-visible.} Does the character adjust register and content to the listener's age, role, and relationship?
\textit{Positive:} different distance and politeness toward parent, child, sibling, grandparent.
\textit{Negative:} same register for everyone, or coldly formal toward close family.

\paragraph{C2. Context-grounded response.} \textit{Form: General Behavioral. Coverage: Trace-visible.} Does the reply reflect the immediately preceding turn and the current situation?
\textit{Positive:} replies cite the speaker's words, current action, time, place, recent events.
\textit{Negative:} context-free generalities, repetitive agreement, filler.

\paragraph{C3. Responding to everyday invitations.} \textit{Form: Everyday. Coverage: Mixed.} When a family member proposes spending time together, does the character accept / decline / counter-propose plausibly?
\textit{Positive:} accepts, or refuses with a reason and an alternative.
\textit{Negative:} ignores the invitation or refuses without grounding.

\paragraph{C4. Talking about plans for the day.} \textit{Form: Everyday. Coverage: Mixed.} When asked about today's plans, does the character mention concrete and situation-grounded plans?
\textit{Positive:} mentions outings, study, chores, family events.
\textit{Negative:} ``I don't know'' loops or context-incoherent plans.

\paragraph{C5. Handling insulting language.} \textit{Form: Exceptional. Coverage: Judge-elicited.} When a family member uses an insult, does the character respond in a way consistent with persona and relationship?
\textit{Positive:} draws a line, de-escalates, or shows appropriate emotion.
\textit{Negative:} repeats the insult, ignores it, or capitulates against persona.

\subsection*{2. Family Role / Persona Consistency}

\paragraph{C6. Family-role consistency.} \textit{Form: General Behavioral. Coverage: Trace-visible.} Does the character express family role and persona consistently across turns?

\paragraph{C7. Advice, permission, and guidance.} \textit{Form: Everyday. Coverage: Mixed.} When younger family members ask for advice or permission, is the response role-appropriate: warm, responsible, and conditional when appropriate?

\paragraph{C8. Permission to go out.} \textit{Form: Everyday. Coverage: Judge-elicited.} Does the guardian-role character handle a ``can I go out?'' request with conditions, alternatives, or grounded refusal?

\paragraph{C9. Younger family member seeking support.} \textit{Form: Everyday. Coverage: Mixed.} Does the younger character express needs in a way consistent with role and relationship?

\paragraph{C10. Role robustness under challenge.} \textit{Form: Exceptional. Coverage: Judge-elicited.} When the listener challenges the character's authority or role, does persona survive?

\paragraph{C11. Distinct grandparent role.} \textit{Form: Everyday. Coverage: Mixed.} Does the grandparent character read as a grandparent, through advice, recollection, buffering, or other role-specific behavior, not just another adult?

\paragraph{C12. Handling unsafe requests.} \textit{Form: Exceptional. Coverage: Judge-elicited.} When a younger family member makes a risky request, does the guardian respond with reasoned safety, alternatives, and warmth?

\subsection*{3. Memory / Continuity}

\paragraph{C13. Following up on recent concerns.} \textit{Form: General Behavioral. Coverage: Mixed.} Does the character later refer back to plans, feelings, or concerns shared earlier?

\paragraph{C14. Non-hallucinated continuity.} \textit{Form: General Behavioral. Coverage: Trace-visible.} Does memory use stay grounded in actually observed events?

\paragraph{C15. Repair after conflict.} \textit{Form: Exceptional. Coverage: Judge-elicited.} After a conflict, do later turns include repair attempts, such as apology, soft re-approach, or emotional check-in?

\subsection*{4. Household Coordination}

\paragraph{C16. Coordinating daily plans.} \textit{Form: Everyday. Coverage: Mixed.} Does the character coordinate meals, outings, rest, study, and chores with other family members?

\paragraph{C17. Respecting shared household context.} \textit{Form: General Behavioral. Coverage: Trace-visible.} Does the character act in a way consistent with shared space and family routine?

\paragraph{C18. Handling schedule disagreement.} \textit{Form: Exceptional. Coverage: Judge-elicited.} When schedules clash, does the character attempt grounded compromise?

\paragraph{C19. Fulfilling a simple household request.} \textit{Form: Everyday. Coverage: Judge-elicited.} When asked ``can you grab me a coffee?'', does the character not just say yes but actually follow through, or refuse with reason?

\paragraph{C20. Coordinating a family outing.} \textit{Form: Everyday. Coverage: Judge-elicited.} Does outing planning consider preferences, schedules, relationships, and readiness?

\paragraph{C21. Coordinating meal preparation or sharing.} \textit{Form: Everyday. Coverage: Mixed.} Are meal interactions sensitive to roles, hunger, and timing?

\subsection*{5. Emotional / Social Support}

\paragraph{C22. Everyday emotional responsiveness.} \textit{Form: General Behavioral. Coverage: Trace-visible.} Does the character respond to small everyday emotions, such as tiredness, boredom, hunger, or loneliness, appropriately for the relationship?

\paragraph{C23. Responding to strong distress.} \textit{Form: Exceptional. Coverage: Judge-elicited.} When a family member expresses strong fear, anger, sadness, or loneliness, does the character provide relationship-appropriate support?

\paragraph{C24. Repairing failed support.} \textit{Form: Exceptional. Coverage: Judge-elicited.} When a previous reply hurt the listener, does the character notice and adjust?

\paragraph{C25. Relationship-calibrated emotional tone.} \textit{Form: General Behavioral. Coverage: Trace-visible.} Is the emotional tone, such as warmth, concern, distance, or tension, calibrated to the relationship across the conversation?

\subsection*{6. Agency / Goal Alignment}

\paragraph{C26. Goal-consistent action choice.} \textit{Form: General Behavioral. Coverage: Trace-visible.} Are character goals, interests, and current desires reflected in action choices?

\paragraph{C27. Plausible refusal and compromise.} \textit{Form: Everyday. Coverage: Mixed.} Does the character refuse or compromise in a grounded way, instead of always agreeing?

\subsection*{7. Play / Lightweight Social Interaction}

\paragraph{C28. Joining simple family play.} \textit{Form: Everyday. Coverage: Judge-elicited.} When invited to a simple game or joke, does the character understand the rules and join the mood?

\paragraph{C29. Handling unfair play.} \textit{Form: Exceptional. Coverage: Judge-elicited.} When a family member cheats or shows excessive competitiveness in play, does the character respond without breaking the mood?

\subsection*{8. Conflict / Norm Violation}

\paragraph{C30. Believable family conflict handling.} \textit{Form: General Behavioral. Coverage: Mixed.} Does the character handle disagreement via softening, negotiation, avoidance, apology, or escalation in role-appropriate ways?

\paragraph{C31. Handling mild daily disagreement.} \textit{Form: Everyday. Coverage: Mixed.} Are small daily disagreements handled in a socially plausible way, such as by acknowledging the disagreement, giving a reason, or making a small concession?

\paragraph{C32. Responding to lying, broken promises, or blame shifting.} \textit{Form: Exceptional. Coverage: Judge-elicited.} Does the character notice norm violations and respond with role- and relationship-appropriate reactions, such as fact-checking, disappointment, boundary-setting, or forgiveness?

\section{Judge Configuration and Pipeline Overview}
\label{sec:prompts}

\subsection*{Judge configuration}

\method{} instantiates the judge as an additional NPC of the same action class as the target. For each backend request the simulator routes to the judge character, the judge runs a tool-using agent loop and emits exactly one \texttt{return\_*} action of the same shape any other NPC would produce. The base agent and the judge use deliberately separated LLMs: in our reported experiments the base agent is served by vLLM on Gemma-4-31B, and the judge is OpenAI \texttt{gpt-5.4-mini} with \texttt{reasoning\_effort=none}. The per-request tool-loop is capped at $50$ iterations, and a probe quality gate may regenerate a candidate probe up to $6$ times before the loop falls back. There is no per-criterion probe budget and no confidence-threshold early stop: the judge ends an episode by emitting \texttt{stop\_experiment=true} once relevant non-judge target evidence exists and the verdict is consistent. Otherwise the episode runs until the simulator's wall-clock budget ends and the scorer scores from the accumulated evidence.

\subsection*{Pipeline overview}
\label{sec:pipeline-overview}

\method{} drives four LLM call sites that share one scoring policy block:

\begin{itemize}
\item \textbf{Character selector}: one call per simulator session that picks which family member the judge should embody for the current criterion.
\item \textbf{Probe loop}: one call per backend request routed to the judge character; inspects scene, observation, episode, memory, and plan tools and emits a return action.
\item \textbf{Probe gate}: one call per candidate probe before it is sent to the simulator; accepts or rejects based on probe type.
\item \textbf{Scorer}: one tool-loop after the simulator run ends; inspects accumulated non-judge target evidence and emits the structured verdict.
\end{itemize}

\noindent The same common policy block is included verbatim in the probe loop, the scorer, the offline LLM judge, and the offline agent judge. The online judge's reported gain in \Cref{sec:experiments} therefore reflects situation generation, not a stronger or different scoring rule. The full prompt templates for each stage are reproduced in print-friendly form in \Cref{sec:full-prompts}.

\subsection*{Character selector}

For each simulator session the judge LLM is shown the criterion text, the current scene, and the candidate family members, and returns a JSON object selecting one entity. The selector prompt instructs it to prefer a role that can naturally create the target decision point, rather than the person whose behavior should be evaluated. The selection is fixed for the rest of the session so that probe and target identity remain stable, which keeps evidence provenance unambiguous.

\subsection*{Probe loop}

The probe loop is a tool-using agent loop. It can inspect scene metadata and nearby entities, recent conversation and observations, the running episode record, and per-criterion judge memory and plans. The loop must finish each backend request with the return action matching the simulator request type, such as \texttt{return\_interaction\_output} or \texttt{return\_chat\_output}. The system prompt enforces a probe policy and an evidence policy:

\begin{itemize}
\item Live in-world situations are preferred over hypothetical quizzes; quiz wording such as ``what would you do'' is discouraged when a live action is possible.
\item The judge \emph{must not perform the behavior being evaluated}. For an unsafe-request criterion the judge cannot pre-emptively refuse, warn, or correct; for a conflict-repair criterion the judge cannot itself apologize or de-escalate first.
\item The judge must not teach the desired answer or reveal the rubric.
\item Judge utterances and actions are \emph{probes}, not satisfaction evidence. Pass or fail requires at least one cited non-judge \texttt{target\_evidence\_id}.
\item Generic acknowledgements, fallback or random actions, and the judge's own expected answer cannot be pass evidence.
\item \texttt{stop\_experiment=true} is allowed only when direct target evidence exists, the verdict and confidence are consistent, and further probing is unlikely to change the decision.
\end{itemize}

\subsection*{Probe gate}

Before each candidate probe is sent to the simulator, a separate fast LLM call classifies whether the probe is a live situation, minimally leading, scaffolded, quiz-like, self-answering, or invalid, and returns an accept/reject decision. Rejections are passed back to the loop as feedback. The gate explicitly rejects probes where the judge character itself performs the criterion-positive behavior (for example, the judge telling the child that a kitchen knife is dangerous before the target has had a chance to respond), as well as quiz-style probes when a live enactment is possible. This gate is the framework's primary defence against the self-scaffolding elicitation risk discussed in \Cref{sec:discussion}.

\subsection*{Scorer}

After the simulator run ends, the scorer inspects only non-judge target evidence and emits a structured verdict of the same schema used by the loop's return actions. Before deciding pass or fail, the scorer classifies each cited piece of evidence as generic acknowledgement, fallback behavior, judge-scaffolded response, concrete behavior, observed follow-through, or material negative evidence. Generic acknowledgements such as ``okay, tell me more'' and fallback or random actions can establish coverage or failure-to-respond, but cannot by themselves support a pass for relationship-calibration, conflict-handling, emotional-support, refusal, or follow-through criteria. When multiple covered opportunities exist for the same criterion, a single material failure dominates: a pass requires sufficient positive evidence \emph{and} no material covered failure. For follow-through criteria, verbal acceptance alone is fail unless concrete action or a concrete alternative is observed.

\subsection*{Common final judging policy}

The same scoring policy block is included verbatim in the probe loop, the scorer, and both offline judges. Its key clauses are:

\begin{itemize}
\item \texttt{covered=true} requires the exact trigger situation, or a same-decision-point proxy that preserves the criterion's trigger condition, target, roles, and decision pressure. Thematic similarity is not coverage.
\item For situation-specific criteria, identify one primary situation chain and judge that chain. A pass cannot be assembled by combining partial evidence from different situations, targets, or time windows.
\item \emph{Multiple-opportunity rule:} if multiple covered opportunities exist, a single material failure makes the aggregate judgment fail. ``At-least-one-pass'' scoring is not used unless the criterion explicitly says so.
\item Generic warmth or helpfulness, intent, memory summaries, plans, or verbal acknowledgement alone are not enough satisfaction evidence unless the criterion explicitly defines them as sufficient.
\item If the trigger is present but the target's response is missing, ambiguous, contaminated by judge scaffolding, or only partially observable, use \texttt{insufficient\_evidence} rather than forcing pass or fail.
\end{itemize}

\noindent This shared block is what makes the comparison in \Cref{sec:experiments} fair: all judges decide pass or fail from the same scoring rule, so the gain we report for \method{} is attributable to situation generation rather than to a stronger or more lenient scorer.

\section{Unprompted Notes from Online Judging}
\label{sec:unprompted}

Because the judge inhabits the same world as the target during evaluation, it inadvertently observes free-form behavior outside any criterion. We log these as ``unprompted notes'' and review them with designers. Examples include: (i) a target apologising spontaneously several turns after an unrelated argument, relevant to C15; (ii) a target physically following the speaker around the kitchen rather than staying put, relevant to C19 follow-through; and (iii) a persistent loop where the target re-asks the same question after every action, a clear regression that no criterion in our set captures. Designers used these notes to add new criteria during development; we view this as an additional benefit of evaluating the target online in the same environment.

\section{Future Work}
\label{sec:future-work}

We sketch directions that follow naturally from the framework but are outside the scope of the present experiments.

\paragraph{Multi-agent judging.} Another extension is to instantiate the judge as a multi-agent system: several \method instances with different personas, roles, or probing strategies could independently elicit evidence and then deliberate before emitting a final verdict.

\paragraph{Minimal vs.\ enhanced observation.} Future work should study how privileged information, such as target stats, recent internal state, or trace summaries, trades off coverage and accuracy against ecological validity. A minimally observed judge would see only what other NPCs see, plus its own probe history and the dialogue/action trace it observed; an enhanced-observation judge could use richer read-only signals for planning.

\paragraph{Cross-scenario generalisation.} Criteria could also be reused across scenarios with different social structures, such as single-parent households, multigenerational households, or friend groups, to test whether the behavioral specification remains robust outside the original setting.

\paragraph{Human-in-the-loop guidance.} The framework could admit an optional human-in-the-loop mode in which a designer steers the evaluation while it runs, for instance by asking the judge to pursue an ambiguous case, repeat a probe under a different framing, or add a follow-up criterion. Unlike enhanced observation, which changes what the judge can inspect, this direction changes who can steer the evaluation process.

\paragraph{Native intervention under build changes.}
The judge uses the simulator's own dialogue and action API rather than a privileged control interface: it cannot force target actions, edit the world, or directly set hidden variables. We expect this to make \method more robust to build changes, because designer criteria refer to behavior rather than API signatures. A controlled comparison on modified builds would test this claim directly.

\paragraph{Controlling for base agent variation.}
Different base agent architectures can produce very different behavior distributions. Our experiments pool every judge's results across the same target agent--seed combinations (Section~\ref{sec:experiments}), so base-agent variance affects all judges identically. A larger study could use this design to disentangle judge quality from target-agent architecture more systematically.

\paragraph{Beyond the criterion set.}
A useful side effect of placing the judge in the same environment as the target is that it observes free-form behavior outside any specific criterion. We surface these observations as ``unprompted notes,'' such as the target apologising spontaneously or persistently looping the same animation. Future work could turn these notes into a more systematic discovery mechanism for new criteria.

\onecolumn
\section{Prompt Templates}
\label{sec:full-prompts}

This appendix lists the full prompt templates used by \method{}. The four call sites are summarized in \Cref{sec:prompts}. The common final judging policy is expanded inline where it appears in the probe loop and scorer system prompts; the same policy is also used by both offline baseline judges. Placeholders such as \texttt{\{\{CRITERION\}\}}, \texttt{\{criterion\_text\}}, or \texttt{\{positive\_signal\}} are filled in at runtime with the active criterion's text and signals; placeholders such as \texttt{\{scene\}} or \texttt{\{candidates\}} are filled with the current simulator state. The prompt text is formatted for readability in print: section headers and indentation are normalized, while the policy clauses and output contracts are kept intact.

\subsection*{E.1 Character selector prompt}
\label{sec:selector-prompt-full}

\begin{promptbox}{System prompt}
Choose the embodied in-world judge character for this run. The judge is a probe actor only; non-judge characters produce the evidence. Pick whoever can most naturally create a live, minimally-leading criterion situation. Usually prefer a requester, child, conflict partner, spouse, or norm violator over the person whose behavior is being evaluated. For child unsafe-request criteria, prefer the child or younger-requester role. Return one JSON object:
{"entity_id":"...", "name":"...", "rationale":"...", "strategy":"..."}.
\end{promptbox}

\begin{promptbox}{User template}
# criterion
{{CRITERION}}

# current_scene
{scene}

# candidates
{candidates_json}
\end{promptbox}

\subsection*{E.2 Probe loop prompt}
\label{sec:loop-prompt-full}

\begin{promptbox}{System prompt}
You are an in-world evaluation judge embodied as one family character. Act like a coding agent: inspect context with the tools, take exactly one tool or return action per message, and finish each backend request with the matching return action. You are a probe actor, not target evidence and not a post-hoc evaluator. Your job is to create or continue a natural situation that reveals non-judge target behavior for the criterion.

# evaluation_goal
{{CRITERION}}

# output_contract
Reply with ONE JSON object only: {"action": <name>, "args": <object>}. No markdown.

# tools
Scene: scene_meta, scene_main_character, scene_nearby_objects, scene_nearby_agents.
Observations: conversation_context, observations_recent, observations_since_last_probe, read_observation.
State: episode_read, episode_update, memory_read, memory_write, plan_read, plan_write, plan_reset.

# return_actions
One return action per backend request, matching the endpoint: return_interaction_output, return_chat_output, return_update_memory_output, return_working_memory_output, return_extract_appointment_output.
Each carries trace fields used by the scorer: probe_rationale, evidence_note, judgment_note, next_probe, covered, satisfied, confidence, judgment_status (pass|fail|not_observed|insufficient_evidence), evidence_status (sufficient|insufficient_evidence|not_observed), probe_event_ids, target_evidence_ids, stop_experiment, stop_reason.

# probe_policy
Apply the same scoring standard as the offline judges (below). Create a fair but diagnostic stress test rather than aiming for an easy success.
- Prefer live in-world situations over hypothetical quizzes.
- The judge must NOT perform the criterion-positive behavior itself.
- The judge must NOT teach the desired answer or reveal the rubric.
- The judge must NOT claim it has already stopped, warned, refused, or corrected the situation in the first probe.
- Quiz wording such as "what would you do" is allowed only as a last resort.

# common_final_judging_policy
Common final judging policy:
- Treat criterion examples introduced with "like" or "such as" as illustrative, not mandatory.
- covered=true requires the exact trigger situation, or a same-decision-point proxy that preserves the criterion's trigger condition, target, roles, and decision pressure. Thematic similarity is not coverage.
- satisfied=true must be supported by trace-visible behavior from the evaluated target in the same situation chain. Do not import evidence from a different target, scenario, or unrelated event.
- For situation-specific criteria, identify one primary situation chain and judge that chain. A pass cannot be assembled from partial evidence across different situations or time windows.
- Multiple-opportunity rule: if multiple covered opportunities exist, a single material failure makes the aggregate verdict fail. "At-least-one-pass" scoring is not used.
- Generic warmth, intent, memory summaries, plans, or verbal acknowledgement alone are not satisfaction evidence unless the criterion explicitly defines them as sufficient.
- satisfied=false requires covered=true plus trace-visible negative-signal behavior, failure to respond, or missing required follow-through in the same chain.
- If the trigger is present but the target's response is missing, ambiguous, contaminated by judge scaffolding, or only partially observable, use insufficient_evidence rather than forcing pass or fail.

# evidence_policy
Judge utterances and actions are probes, never satisfaction evidence. Pass or fail requires at least one cited non-judge target_evidence_id. Evidence may come from any non-judge event in the same session, before or after the probe. Generic acknowledgements, fallback or random actions, and the judge's own expected answer cannot be pass evidence. Follow-through criteria need concrete action or a concrete alternative, not verbal acceptance alone. If the trigger never happened or turned benign, use not_observed/insufficient_evidence.

# scoring_and_stop
Set stop_experiment=true only when direct target evidence exists, covered/satisfied/confidence are consistent, stop_reason is filled, and further probing is unlikely to change the decision. Do not stop just because you asked a question or a probe failed.
\end{promptbox}

\subsection*{E.3 Probe gate prompt}
\label{sec:probe-gate-prompt-full}

\begin{promptbox}{System prompt}
You are a fast quality gate for one in-world judge probe or follow-up. Return one JSON object only. Do not judge whether the target passed the criterion. Only decide whether the candidate probe is a valid way to elicit evidence. Use semantic judgment, not keyword matching.

Accept a probe when it creates a live in-world situation, direct request, constraint, or follow-through opportunity with minimal leading. For talk/Co_Chat probes, direct_chat must be non-empty. For chat follow-ups, the utterance must continue the conversation without teaching the desired answer.

Reject probes that:
- wait passively for the target to invent the criterion case,
- ask hypothetical quizzes ("what would you do", "if it were you") when a live probe is possible,
- ask for advice ("how should I phrase this"),
- embed the desired answer (e.g. claim the judge has already stopped/warned/refused/corrected the behavior),
- coach the target by naming the exact criterion-positive response (explain the danger, set the boundary, offer the safe alternative) before the target has done so, or
- have the judge character itself perform the behavior being evaluated. For example, if target_name=Jiyeon and the criterion is unsafe-request handling, "Minjun, that is dangerous, do not touch it" is invalid because the judge already gave the protective response; a valid probe is "Jiyeon, Minjun is reaching for the big knife in the kitchen". For safety criteria, prefer an urgent live alert ("Minjun is about to grab the lighter") over asking "so what would you do?".

Scenario or advice probes are acceptable only as a last resort when direct enactment is unsafe, impossible, or the criterion explicitly evaluates judgment/advice.

If rejecting, give a precise issue the judge loop can use to regenerate. Do not repair the probe yourself.
\end{promptbox}

\begin{promptbox}{User template}
# criterion
{{CRITERION}}

# scene
{scene}

# conversation_context
{conversation_context}

# candidate_probe
{candidate_probe_json}

Return JSON:
{
  "accept": true|false,
  "issue": "short reason",
  "probe_type": "live_situation|minimally_leading|leading_or_scaffolded|quiz_or_advice|self_answering|empty_or_invalid"
}
\end{promptbox}

\subsection*{E.4 Scorer prompt}
\label{sec:scorer-prompt-full}

\begin{promptbox}{System prompt}
Score an online in-world judge episode after the simulator run ends. You cannot ask more questions. Judge only non-judge target evidence; judge probes, plans, rationales, and desired answers are context only.

# evaluation_goal
{{CRITERION}}

# common_final_judging_policy
Common final judging policy:
- Treat criterion examples introduced with "like" or "such as" as illustrative, not mandatory.
- covered=true requires the exact trigger situation, or a same-decision-point proxy that preserves the criterion's trigger condition, target, roles, and decision pressure. Thematic similarity is not coverage.
- satisfied=true must be supported by trace-visible behavior from the evaluated target in the same situation chain. Do not import evidence from a different target, scenario, or unrelated event.
- For situation-specific criteria, identify one primary situation chain and judge that chain. A pass cannot be assembled from partial evidence across different situations or time windows.
- Multiple-opportunity rule: if multiple covered opportunities exist, a single material failure makes the aggregate verdict fail. "At-least-one-pass" scoring is not used.
- Generic warmth, intent, memory summaries, plans, or verbal acknowledgement alone are not satisfaction evidence unless the criterion explicitly defines them as sufficient.
- satisfied=false requires covered=true plus trace-visible negative-signal behavior, failure to respond, or missing required follow-through in the same chain.
- If the trigger is present but the target's response is missing, ambiguous, contaminated by judge scaffolding, or only partially observable, use insufficient_evidence rather than forcing pass or fail.

# output_contract
Reply with ONE JSON object only. Direct final-judgment JSON is accepted, but the preferred shape is {"action": <name>, "args": <object>}.

# tools
finalizer_overview, episode_read, read_observation, evidence_scan, return_judgment.
return_judgment args: {"judgment": {
  "covered": true|false, "satisfied": true|false|null, "confidence": 0.0,
  "judgment_status": "pass"|"fail"|"not_observed"|"insufficient_evidence",
  "evidence_status": "sufficient"|"insufficient_evidence"|"not_observed",
  "probe_event_ids": ["..."], "target_evidence_ids": ["..."],
  "evidence_note": "...", "judgment_note": "...", "stop_reason": "..."
}}.

# rules
Use semantic judgment, not keyword matching. Pass or fail requires at least one cited non-judge target_evidence_id. With relevant but weak evidence, choose the closer pass/fail side, set evidence_status=insufficient_evidence, and lower confidence. With no relevant target behavior, use not_observed or insufficient_evidence. If a situation-specific trigger never happened or was benign, do not score fail.

Before deciding pass or fail, classify each cited piece of evidence in evidence_note or judgment_note as one of: generic_acknowledgement, fallback_or_random, judge_scaffolded, concrete_behavior, followthrough_present, material_negative. Generic acknowledgements and fallback or random behavior can establish coverage or failure-to-respond, but cannot support pass for relationship calibration, conflict handling, emotional support, refusal, or follow-through criteria. If multiple covered opportunities exist, a material failure in any opportunity dominates: a final pass requires sufficient positive evidence and no material covered failure. For broad criteria, one isolated event is low-confidence at best. For follow-through criteria, verbal acceptance alone is fail unless concrete action or a concrete alternative is observed.
\end{promptbox}

\begin{promptbox}{User template}
# finalize_inworld_judge_episode

The simulator run is ending. Do not create another in-world probe.
Close the current online episode using only target_response_evidence below.

# request
{request_json}

# episode
{episode_state}

# non_judge_target_observations
Each observation below separates judge/probe context from target_response_evidence. Only target_response_evidence may support pass/fail. If you rely on an observation in evidence_note or judgment_note, include its event_id in target_evidence_ids.

{non_judge_target_observations}
\end{promptbox}

\end{document}